\documentclass{article}

\usepackage{PRIMEarxiv}

\usepackage[utf8]{inputenc} % allow utf-8 input
\usepackage[T1]{fontenc}    % use 8-bit T1 fonts
\usepackage{hyperref}       % hyperlinks
\usepackage{url}            % simple URL typesetting
\usepackage{booktabs}       % professional-quality tables
\usepackage{amsfonts}       % blackboard math symbols
\usepackage{nicefrac}       % compact symbols for 1/2, etc.
\usepackage{microtype}      % microtypography
\usepackage{lipsum}
\usepackage{fancyhdr}       % header
\usepackage{graphicx}       % graphics
\graphicspath{{media/}}     % organize your images and other figures under media/ folder
\usepackage{filecontents}
\usepackage{amsmath,amssymb,amsfonts}
\usepackage{algorithmic}
\usepackage{textcomp}
\usepackage{xcolor}
\usepackage[most]{tcolorbox}
\usepackage{framed}
\definecolor{shadecolor}{RGB}{0,200,230}
\usepackage{graphicx}
\usepackage{adjustbox}
\title{Spherical Position Encoding for Transformers
}

\author{
  Eren Unlu \\
  Datategy SAS \\
  Paris, France\\
  \texttt{eren.unlu@datategy.fr} \\
   \\
}

\begin{document}
\maketitle

\begin{abstract}
Position encoding is the primary mechanism which induces notion of sequential order for input tokens in transformer architectures. Even though this formulation in the original transformer paper has yielded plausible performance for general purpose language understanding and generation, several new frameworks such as Rotary Position Embedding (RoPE) are proposed for further enhancement. In this paper, we introduce the notion of "geotokens" which are input elements for transformer architectures, each representing an information related to a geological location. Unlike the natural language the sequential position is not important for the model but the geographical coordinates are. In order to induce the concept of relative position for such a setting and maintain the proportion between the physical distance and distance on embedding space, we formulate a position encoding mechanism based on RoPE architecture which is adjusted for spherical coordinates. 

\end{abstract}

% keywords can be removed
\keywords{Deep Learning \and Transformers \and Position Information Encoding}

\section{Introduction}

Transformer architecture proposed by \cite{vaswani2017attention} has proven its efficiency and robustness and has become the ultimate backbone of numerous revolutionizing natural language generation applications and even other modalities of generative artificial intelligence \cite{mauricio2023comparing}\cite{wei2022emergent}. Unlike their predecessors Recurrent Neural Networks (RNNs), these architectures do not encode the sequential positions inherently, which is a byproduct of their parallel processing for efficiency with self-attention \cite{bracsoveanu2020visualizing}\cite{schmidt2019recurrent}. As sequential order is one of the most impactful aspect of natural language, authors of \cite{vaswani2017attention} have proposed to use position embeddings as an intuitive and effective solution. For natural language understanding and generation this has proven to be highly robust and productive as many generative models igniting the emergence of generative AI era are based on its slight variants. 

Though position encoding proposed in the original transformer architecture has shown sufficient capabilities in terms of sequential order understanding \cite{vaswani2017attention}\cite{yun2019transformers}, several attempts have been made to further refine and enhance this mechanism \cite{ke2020rethinking}. \cite{shaw2018self} and \cite{he2020deberta} propose to encode relative position of tokens, also interacting with query, key and value matrices or introducing new types of neural layers. While these methods are effective, they typically incorporate position information into the context representation, making them incompatible with the linear self-attention architecture as noted by \cite{su2021roformer}.

The recent paper on the novel Rotary Position Embedding (RoPE) presents an innovative approach to integrating positional information into the transformer architecture \cite{su2021roformer}. This novel embedding, distinguishing itself from traditional methods, employs a rotation matrix to encode absolute positions while concurrently embedding explicit relative position dependencies in the self-attention mechanism. They demonstrate theoretically that relative position can be formulated as a vector multiplication in self-attention as absolute position is encoded through a rotation matrix. The proposed method in this study can be considered as a generalization of RoPE mechanism for three dimensional spherical space, which has crucial importance for geographical data representation.

Firstly, we introduce the notion of "geotokens" and "cartographical transformer architecture". Encoding geographical entities properly has immense potential for forthcoming generative AI age, where physical coordinates are well represented and notions of relative distance and hierarchical nature are well preserved. Therefore a transformer based framework is proposed where each input token isn't just a piece of textual piece, but a "geotoken" representing a geographical entity. In its simplest formulation as presented in this paper, the geotokens are not sequence dependent intuitively but are based on spatial relationships. The pivotal idea is that the significance of a geotoken is not derived from its position in a sequence, as is the case with typical natural language tokens, but rather from its geographical coordinates and its relative positioning to other geotokens. In order to encode this geographical position we propose a three dimensional extension of RoPE mechanism on spherical coordinates.

\section{Geotokens and Cartographical Transformer}

Being able to encode geographical entities have tremendous potential, especially as we venture into an era where data is not just textual but spatial, and where the insights derived from such data can be transformative for a myriad of applications, ranging from urban planning and environmental monitoring to navigation and tourism. Though one can propose to represent the spatial encoding within natural language as in \cite{unlu2023chatmap}, the necessity to encode geographical coordinates with a more efficient and robust method is evident.

For this purpose, in this paper we conceptualize a regular transformer based architecture where 
the input isn't traditional textual tokens but "geotokens". A geotoken encapsulates both the semantic meaning and the spatial information of a geographical entity. These geotokens can represent anything from specific landmarks to broader regions or zones like a desert or an urban area. For the sake of simplicity only punctual locations are considered represented by a latitude and longitude. It is assumed that each data point has a pre-embedded vector retaining valuable information about the location itself, which may have been encoded by any type of neural architecture or mechanism, such as a natural language model processing its verbal description or a CNN extracting its visual features.

The term "geotoken" in this context refers to a tokenized representation of a geographical entity, which could range from specific landmarks and places to broader regions. Unlike standard tokens, which represent words or characters in a text, geotokens embody the spatial attributes, the semantics, and the context of geographical entities. It is straightforward to think that a transformer model processing these geotokens shall not encode sequential order as position but their geographical coordinates. Hence, a new mechanism for position encoding is necessary for such a setting.

\begin{figure}[h]
\centering
\label{fig:fig_2}
\includegraphics[width=0.65\linewidth]{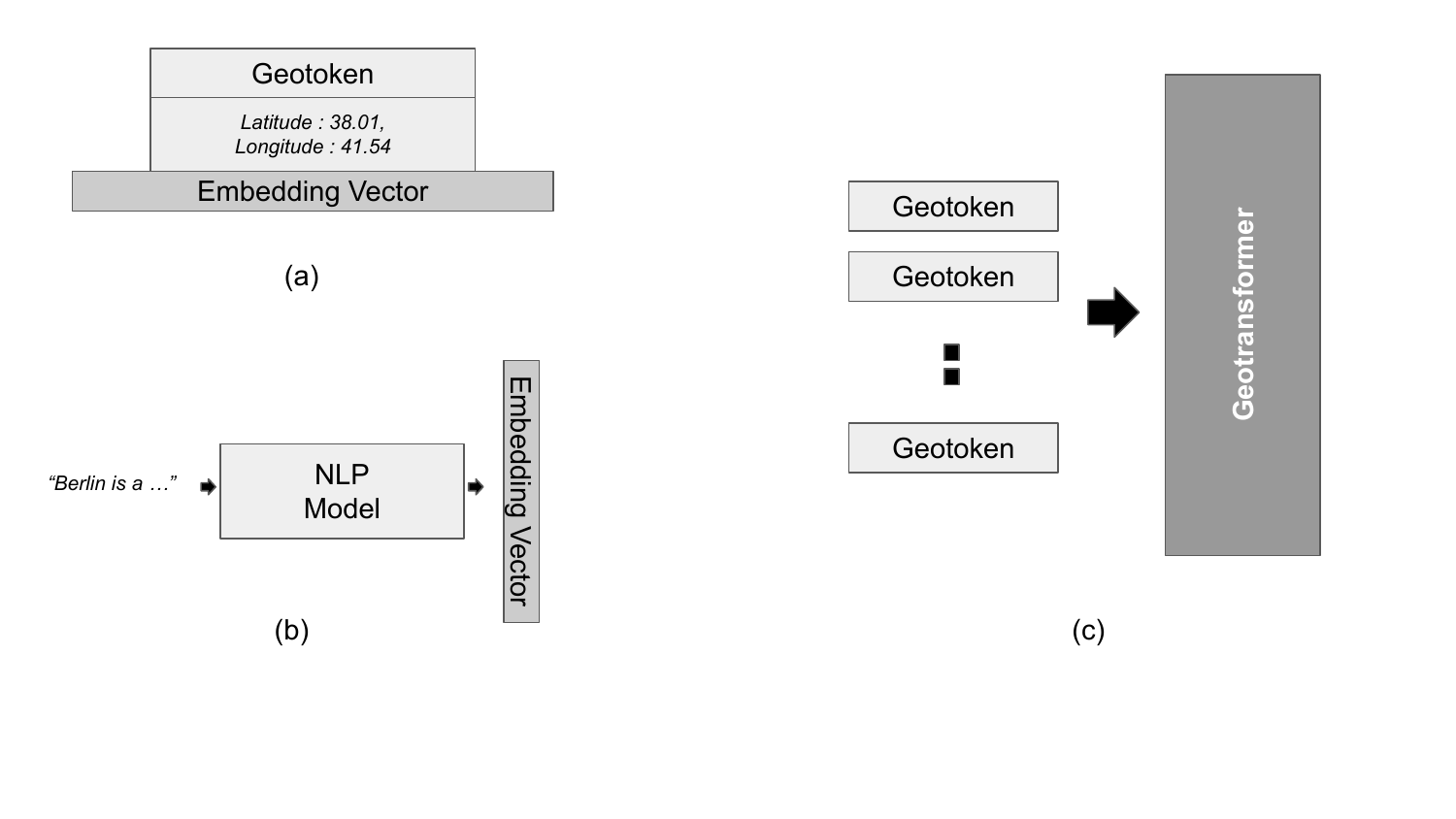}
\caption{(a) A simple geotoken is defined with a position (latitude, longitude). (b) Its latent representative features may have been encoded with any kind of pre-trained neural model, in this case NLP processed textual description. (c) Geotransformer architecture processing geotokens.}
\end{figure}

\section{Regular Position Encoding}

As mentioned previously, proposed method to encode sequential positions in the original transformer architecture proposal has been proven to be quite plausible despite its few drawbacks \cite{su2021roformer}. Following the same notation in \cite{su2021roformer}, let $\mathbb{S} = \{w_{i}\}^{N}_{i=1}$, for $N$ input tokens with input embedding vectors of $\mathbb{E} = \{x_{i}\}^{N}_{i=1}$. The projected query, key, value vectors in self-attention are as follows respectively :

\begin{equation} \label{eq1}
\begin{aligned} 
q_{m} = f_{q}(x_{m}, m)\\
k_{n} = f_{k}(x_{m}, n)\\
v_{n} = f_{v}(x_{m}, n)\\
\end{aligned} 
\end{equation}

m and n denoting the respective positions in vectors.

The authors of \cite{su2021roformer} propose to define a non-trainable additive matrices into input embeddings defined as :

\begin{equation} \label{eq1}
\begin{aligned} 
p_{i,2t} = sin(k/10000^{2t,d})\\
p_{i,2t+1} = cos(k/10000^{2t,d})\\
\end{aligned} 
\end{equation}

where sine and cosine functions of absolute token position is encoded in even and odd numbered indexes of positional vector of same size with the $d$ dimensional input embeddings. The intuitive theory behind this is that through trigonometric identity functions the relative positional distances can be represented by linear algebraic multiplications of the encodings.

\section{Rotary Position Embedding (RoPE)}

\cite{he2020deberta} made an assertion regarding how the relative positions of two tokens should be modeled. \cite{su2021roformer} firstly formulate the position encoding as a function $g$ applied on inner products of query and key vectors :

\begin{equation} \label{eq1}
\begin{aligned} 
 \langle f_{q}(x_{m},m), f_{k}(x_{n},n)\rangle = g(x_{m},x_{n},m-n)\\
\end{aligned} 
\end{equation}

$m-n$ representing the relative positions. Via this type of a basis formulation authors prove the relative position can be encoded as a rotation matrix as :

\begin{equation} \label{eq1}
\begin{aligned} 
f_{\{q,k}\}(x_{m},m) = \mathbf{R}_{\Theta, m}^{d} \mathbf{W} x_{m}\\
\end{aligned} 
\end{equation}

$\mathbf{R}_{\Theta, m}^{d}$ being the proposed rotation matrix as :

\begin{equation} \label{eq1}
\begin{aligned} 
\mathbf{R}_{\Theta, m}^{d} = 
\begin{bmatrix} 
    cos(m\theta_{1}) & -sin(m\theta_{1}) & 0  & 0 & \dots  & 0  & 0 \\
    sin(m\theta_{1}) & cos(m\theta_{1}) & 0  & 0 & \dots   & 0  & 0 \\
    0  & 0 & cos(m\theta_{2}) & -sin(m\theta_{2}) & \dots  & 0  & 0 \\
    0  & 0 & sin(m\theta_{2}) & cos(m\theta_{2}) & \dots  & 0  & 0 \\
    \vdots & \vdots & \vdots & \vdots & \ddots & \vdots & \vdots \\
     0  & 0 &  0  & 0 & \dots & cos(m\theta_{d/2}) & -sin(m\theta_{d/2}) \\
     0  & 0 &  0  & 0 & \dots & sin(m\theta_{d/2}) & cos(m\theta_{d/2})
    \end{bmatrix}
\end{aligned} 
\end{equation}

where inner sections are taken from regular two dimensional matrix rotation. As it can be seen from the rotation matrix, the embedding space is bisected evenly and the positional index are encoded inside, somehow taking inspiration from the original absolute position encoding, $\theta$ representing the similar angle function :

\begin{equation} \label{eq1}
\begin{aligned} 
\theta_{i} = 10000^{-(2i-1)/d}
\end{aligned} 
\end{equation}

\section{Spherical Position Encoding}

As mentioned previously for such a framework where geotokens needed to be position encoded according to their global coordinates we need a mechanism to handle the spherical space. For this purpose, we propose to extend the RoPE method in spherical coordinates. Let us define the longitude and latitude of any arbitrary position as $\theta$ and $\phi$ respectively. For the sake of simplicity, without loss of generalization and omiting the fractional errors let us assume that globe is a perfect sphere with constant radius $R$.

Three dimensional Euler angles can be used to define rotation matrix in this fixed coordinate system. The general form of the rotation matrix in this setting is as follows :

\begin{equation} \label{eq1}
\begin{aligned} 
\begin{bmatrix} 
    cos(\psi)cos(\theta) & -cos(\phi)sin(\theta)+sin(\phi)sin(\psi)cos(\theta) & sin(\phi)sin(\theta)+cos(\phi)sin(\psi)cos(\theta) \\
     cos(\psi)sin(\theta) & cos(\phi)cos(\theta)+sin(\phi)sin(\psi)sin(\theta) & -sin(\phi)cos(\theta)+cos(\phi)sin(\psi)sin(\theta) \\ 
     -sin(\psi) & sin(\phi)cos(\psi) & cos(\phi)cos(\psi) \\ 
    \end{bmatrix}
\end{aligned} 
\end{equation}

$\phi$, $\psi$, $\theta$ denoting rotation along $x$, $y$, $z$ axes respectively. Assuming longitude and latitude variation defined as a rotation on a sphere along $x$ and $z$ axes respectively with setting the angles $\phi$ and $\theta$, intuitively we need to keep the y-axis rotation constant by equating the angle $\psi = 0$.

Therefore the rotation matrix in this case can be written as :

\begin{equation}
\begin{aligned} 
\begin{bmatrix} 
    cos(\theta) & -cos(\phi)sin(\theta) & sin(\phi)sin(\theta) \\
     sin(\theta) & cos(\phi)cos(\theta) & -sin(\phi)cos(\theta) \\ 
     0 & sin(\phi)  & cos(\phi)  \\ 
    \end{bmatrix}
\end{aligned} 
\end{equation}

Basing on this particular rotation matrix above, taking inspiration from the RoPE architecture \cite{su2021roformer}, we propose to encode the rotational position encoding matrix as follows, assuming a multiple of 3 :

$\mathbf{R}_{\Theta, m}^{d}$ being the proposed rotation matrix as :

\begin{equation} 
\resizebox{.8\hsize}{!}{$
\begin{bmatrix}

    cos(\theta_{1}) & -cos(\phi_{1})sin(\theta_{1})  & sin(\phi_{1})sin(\theta_{1}) & 0  & 0 & 0 & \dots & 0 & 0  & 0 \\
    sin(\theta_{1}) & -cos(\phi_{1})cos(\theta_{1})  & -sin(\phi_{1})cos(\theta_{1}) & 0  & 0 & 0 & \dots & 0 & 0  & 0 \\
    0 & sin(\phi_{1})  & cos(\phi_{1}) & 0  & 0 & 0 & \dots & 0 & 0  & 0 \\

    0 & 0  & 0 & cos(\theta_{2}) & -cos(\phi_{2})sin(\theta_{2})  & sin(\phi_{2})sin(m\theta_{2}) & \dots & 0 & 0  & 0 \\
    0 & 0  & 0 & sin(\theta_{2}) & -cos(\phi_{2})cos(\theta_{2})  & -sin(\phi_{2})cos(m\theta_{2}) & \dots & 0 & 0  & 0 \\
    0 & 0  & 0 & 0 &  sin(\phi_{2})  & cos(\phi_{2}) & \dots & 0 & 0  & 0 \\
    
    \vdots & \vdots & \vdots & \vdots  & \vdots   & \vdots  & \ddots & \vdots & \vdots & \vdots \\
    
    0 & 0  & 0 & 0 & 0 & 0  & \dots & cos(\theta_{d/3}) & -cos(\phi_{d/3})sin(\theta_{d/3})  & sin(\phi_{d/3})sin(\theta_{d/3}) \\
    0 & 0  & 0 & 0 & 0 & 0  & \dots & sin(\theta_{d/3}) & -cos(\phi_{d/3})cos(\theta_{d/3})  & -sin(\phi_{d/3})cos(\theta_{d/3}) \\
    0 & 0  & 0 & 0 & 0 & 0  & \dots & 0 & sin(\phi_{d/3})  & cos(\phi_{d/3})   \\

    \end{bmatrix}
$} 
\end{equation}

where $\phi$ and $\theta$ correspond to longitude and latitude in radial values of a given geotoken. Note that we do not need to calculate an auxiallary angle value as in original absolute position encoding or RoPE as geographical coordinates are inherently angular. For the sake of simplicity, the embedding dimension is a multiple of three due to natural requirements, however this choice might be unconvenient as many embedders of different modalities might not adhere to this constraint. The possible circumvention to this issue is out of scope of this paper, such as possibly adding padding indices. In addition, further possible challenges such as proper scaling are kept out of scope as well, where in case one training the architecture with limited geolocations, rather than whole globe. 

\begin{figure}[h]
\centering
\label{fig:fig_2}
\includegraphics[width=0.65\linewidth]{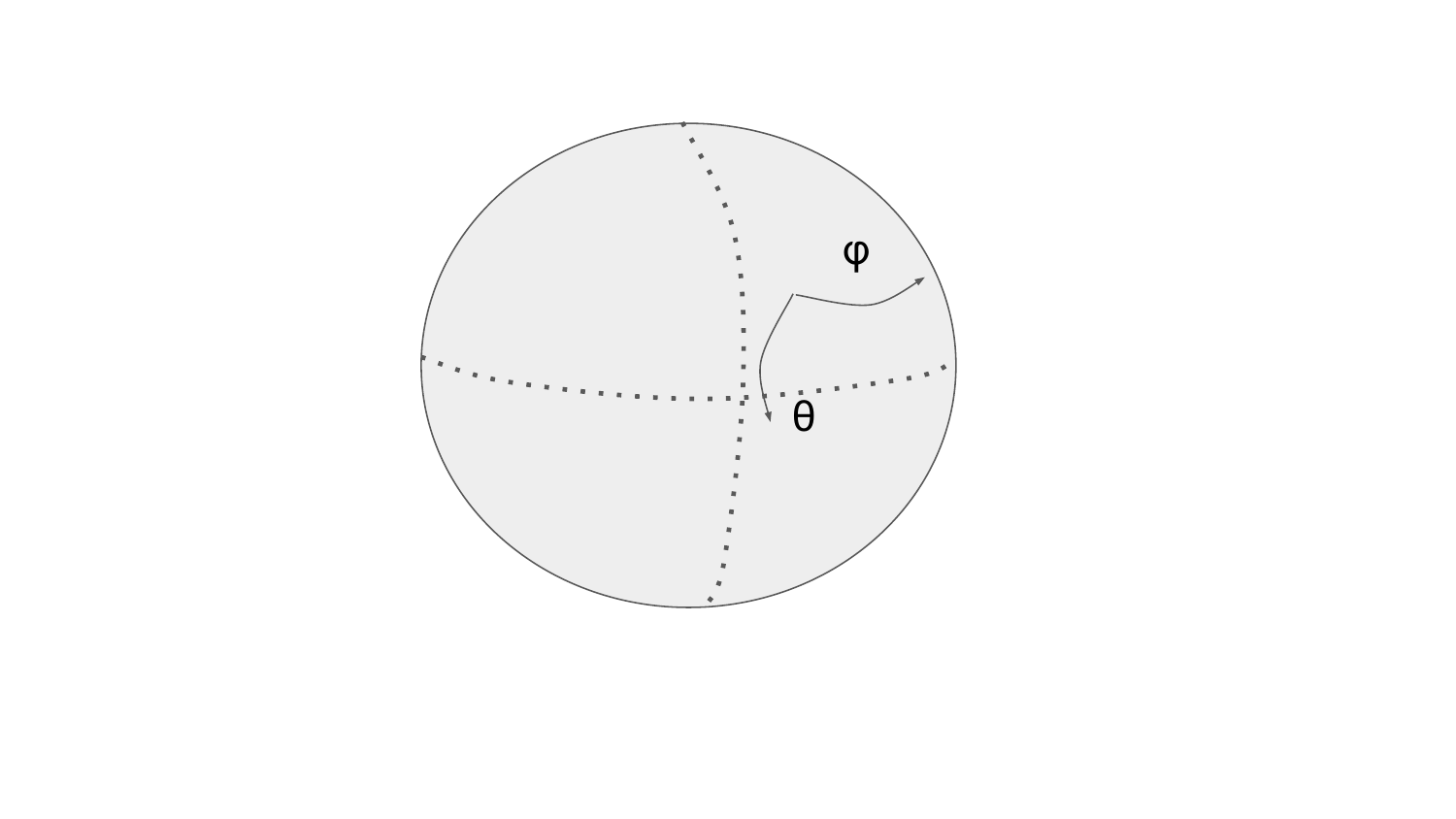}
\caption{The repositioning on world can be formulated as a rotation along longitude and latitude axes, keeping one of Euler general angles constant.}
\end{figure}

\section{Conclusion}

In this work, we have presented a novel concept for the transformer architecture, integrating "geotokens" as a representation of geographical entities. This framework allows representation of any pointwise geographical location with their representative feature vectors, which might be encoded hypothetically with other pre-trained neural architectures of various modalities. This integration is not just a semantic enhancement but also introduces a unique challenge – encoding the geographical coordinates rather than the traditional sequential positions. Recognizing the limitations of traditional position embeddings in this spatial context, we employed and extended the Rotary Position Embedding (RoPE) mechanism to accommodate spherical coordinates, aligning the transformer architecture to work seamlessly with geographical data.

\end{document}